\documentclass[twocolumn,10pt,letterpaper]{article}

\usepackage{cogsci}
\usepackage{comment}
\cogscifinalcopy % Uncomment this line for the final submission 
\usepackage[table,xcdraw,dvipsnames]{xcolor}
\definecolor{citecolor}{HTML}{1F3B4D}

\usepackage{hyperref}
\hypersetup{
    unicode,                    % Use unicode for links
    pdfborder       = {0 0 0},  % Suppress border around pdf
    bookmarksdepth  = subsection,
    bookmarksopen   = true,     % Expand the bookmarks as soon as the pdf file is opened
    breaklinks      = true,
    colorlinks      = true,
    linkcolor       = citecolor,
    citecolor       = citecolor,
    urlcolor        = citecolor,
}

\usepackage{amsmath}
\usepackage{amssymb}
\usepackage{bm}
\usepackage{mathtools}

\usepackage{enumitem}

\usepackage{caption}
\usepackage{subcaption}
\usepackage{graphicx}
\usepackage{dblfloatfix}
\usepackage{multirow}
\usepackage{wrapfig,lipsum,booktabs}
\usepackage{array}
\usepackage{rotating} % for sideways tables
\usepackage{float} % Roger Levy added this and changed figure/table
\usepackage{tikz}

\usepackage{tabularx}

\definecolor{myorange}{RGB}{255,165,0}
\definecolor{myblue}{RGB}{0,0,128}
\definecolor{mygreen}{RGB}{0,128,0}
\definecolor{myred}{RGB}{255,10,10}
\definecolor{myviolet}{RGB}{138,43,226}

\definecolor{observed}{RGB}{170,170,170} 
\definecolor{inference}{RGB}{255,91,89}  
\definecolor{gptfouro}{HTML}{08306B}
\definecolor{claude}{HTML}{66C2A5} 
\definecolor{gemini}{HTML}{A6D854}
\definecolor{humans}{HTML}{E7298A}
\definecolor{gptthreefive}{HTML}{6BAED6}

\DeclareRobustCommand{\colordot}[1]{\begin{tikzpicture}[baseline=(a.south)]
    \node[circle, scale=0.75,color=black, fill=#1] (a) {};
\end{tikzpicture}}

\DeclareRobustCommand{\dashedcircle}{%
    \begin{tikzpicture}[baseline=(a.south)]% Align baseline to the center
        \node[circle, scale=0.75, draw=black, dashed, dash pattern=on 1pt off 1pt, fill=white] (a) {};
    \end{tikzpicture}%
}

\DeclareRobustCommand{\colorsquare}[1]{\begin{tikzpicture}[baseline=(a.south)]
    \node[rectangle, scale=0.75, color=black, fill=#1, minimum width=0.6em, minimum height=0.6em] (a) {};
\end{tikzpicture}}

\DeclareRobustCommand{\colordotcircum}[1]{\begin{tikzpicture}[baseline=(a.south)]
    \node[circle, scale=0.75,color=black, fill=white] (a) {};
\end{tikzpicture}}

\usepackage{pslatex}
\usepackage{apacite}

\usepackage[
    capitalize,
    nameinlink,
]{cleveref}

\title{Do Large Language Models Reason Causally Like Us? Even Better?}

\author{
    {\large \bf Hanna M. Dettki} \\ New York University\\ \href{mailto:hmd8142@nyu.edu}{hmd8142@nyu.edu}
    \And 
    {\large \bf Brenden M. Lake } \\ New York University \\ \href{mailto:brenden@nyu.edu}{brenden@nyu.edu}
    \And 
    {\large \bf Charley M. Wu }\\ University of Tübingen \\ \href{mailto:charleymswu@gmail.com}{charley.wu@uni-tuebingen.de}
    \And
    {\large \bf Bob Rehder } \\ New York University \\ \href{mailto:bob.rehder@nyu.edu}{bob.rehder@nyu.edu} 
}
\begin{document}

\maketitle

\begin{abstract}

Causal reasoning is a core component of intelligence. Large language models (LLMs) have shown impressive capabilities in generating human-like text, raising questions about whether their responses reflect true understanding or statistical patterns. We compared causal reasoning in humans and four LLMs using tasks based on collider graphs, rating the likelihood of a query variable occurring given evidence from other variables. 
LLMs’ causal inferences ranged from often nonsensical (GPT-3.5) to human-like to often more normatively aligned than those of humans (GPT-4o, Gemini-Pro, and Claude). Computational model fitting showed that one reason for GPT-4o, Gemini-Pro, and Claude's superior performance is they didn't exhibit the ``associative bias'' that plagues human causal reasoning. Nevertheless, even these LLMs did not fully capture subtler reasoning patterns associated with collider graphs, such as ``explaining away''. 
%We found that the LLMs' causal inferences varied along a spectrum from often nonsensical (GPT-3.5) to ones that corresponded to a normative account of causal reasoning more closely than the inferences drawn by humans (GPT-4o, Gemini-Pro and Claude). Computational model fitting revealed that one reason for GPT-4o, Gemini-Pro and Claude's superior performance is they didn't exhibit the ``associative bias'' that plagues human casual reasoning. Nevertheless, even these LLMs did not fully exhibit some of the more subtle reasoning patterns associated with collider graphs, such as ``explaining away.''
%to human-like to normative inference, with alignment shifting based on model, context, and task.
%Overall, GPT-4o, Gemini-Pro and Claude showed the most normative behavior, whereas  GPT-3.5 responses were generally poor.
%including partial ``explaining away'', 
%Although some agents deviated from the expected independence of causes  they exhibited strong predictive inference when assessing the likelihood of the effect given its causes.
These findings underscore the need to assess AI biases as they increasingly assist human decision-making.

  \end{abstract}

%%%%%%%%%%%%%%%%%%%%%%%%%%%%%%%%%%%%%%%%%%%%%%%%%%%%%%%%%%%
\section{Introduction}
%%%%%%%%%%%%%%%%%%%%%%%%%%%%%%%%%%%%%%%%%%%%%%%%%%%%%%%%%%%

% the remarkable capabilities of LLMs
Large Language Models (LLMs) have proven to be highly capable across a range of domains, including natural language understanding, answering questions, and engaging in creative tasks \cite{bubeck2023sparks,abdin2024phi,gunter2024apple}.
% are we entering an era of AGI?
In light of these recent advancements in LLMs, many believe that we are now truly entering an era of Artificial Intelligence \cite<AI;>[]{bottou2023borges}. The degree to which machines genuinely comprehend our environment carries significant implications for their reliability in various domains \cite{mitchell2023debate}, including the automatic generation of news content, policy recommendations \cite{kekiC2023evaluating}, 
knowledge discovery, disease diagnosis \cite{nori2023capabilities}, and autonomous driving. 
The impressive capability of LLMs to produce text resembling human language raises the question of whether these models possess some form of world understanding, and if they reason similarly to humans.

% Causal reasoning has a hallmark of intelligence
\emph{Causal reasoning} is widely regarded as a core aspect of  intelligence \cite{lake2017building}. It involves recognizing and inferring the causal relationships between variables, moving beyond mere correlations to uncover underlying mechanisms. Such capabilities are essential in practical applications, including the development of pharmaceutical drugs or the planning of public health strategies. Therefore, causal reasoning is considered an important milestone in the pursuit of Artificial General Intelligence \cite<AGI;>[]{obaid2023machine}.
Causal reasoning can be formalized using \textit{causal Bayes nets} (CBNs) providing a probabilistic calculus for reasoning about the probability of some variables given others that are causally related \cite{pearl1995bayesian}.
By comparing human reasoners to CBNs, CBNs can  serve as a normative benchmark \cite{glymour2003learning, waldmann2006beyond} and help reveal human biases that deviate from ideal causal reasoning \cite{rehder2017failures, bramley2015conservative}. 
% CBNS: a normative benchmark on how to evaluate human reasoning
% Biases in Human Reasoning
For instance, when reasoning about a simple collider graph $C_1 \rightarrow E \leftarrow C_2$, people exhibit biases such as \textit{weak explaining away} and \textit{Markov violations}  \cite<explained later;>[]{rehder2017failures}. These systematic deviations highlight the interplay between normative principles and cognitive heuristics in human causal reasoning.

% Concern: 
% Do LLMs rely on patterns in data?
A plethora of recent studies have assessed the capabilities of LLMs \cite<e.g.,>[]{kiciman2023causal}, 
and concerns have been raised regarding their reliance on learned patterns rather than genuine causal relationships \cite{willig2023causal,jiang2024peek}. For example, \citeA{pmlr-v202-shi23a} and \citeA{mirzadeh2024gsm} demonstrated that introducing irrelevant context can drastically alter the outputs of LLMs. 
That even minor distractions influence their responses raises questions about the robustness of LLMs in high-stakes scenarios.

Indeed, a growing number of researchers have proposed that current LLMs are unable to generalize causal ideas beyond their training distribution and/or without strong user-induced guidance  \cite<e.g., chain-of-thought prompting;>[]{jin2023cladder, kiciman2023causal}. 
Thus, understanding the extent to which LLMs reason causally, and whether they show similar biases to people when they deviate from normative principles has practical importance in deploying AI systems.

  \begin{figure}[t]
    \includegraphics[width=\columnwidth]{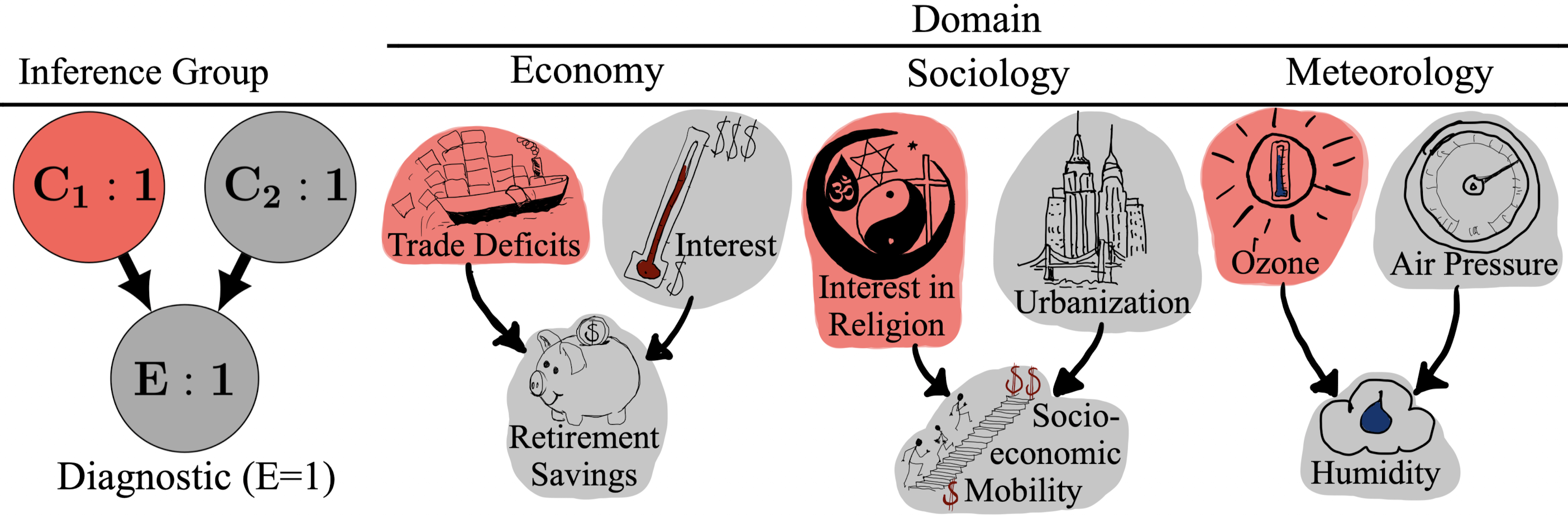}
    \caption{\textbf{Visualization of Causal Mechanism per Domain.} The left most graph represents task VI from the diagnostic inference group.  The  nodes are colored according to:  \colordot{inference} $\to$ latent (query node); \colordot{observed} $\to$ observed $\in \{0,1\}$.
    %; and \dashedcircle{}   $\to$  no information on.} 
    }
    \label{fig:example_table}
  \end{figure}

To this end, \citeA{jin2023cladder} introduced the CLADDER dataset, comprising
10,000 causal reasoning questions designed to evaluate the formal causal reasoning abilities of LLMs. While they tested colliders in their dataset, they didn't contrast LLMs with humans. In addition, although the dataset serves as a valuable benchmark for assessing whether LLMs honor probabilistic rules, solving its tasks requires substantial background knowledge (college-level statistics and pen and paper), making it less suitable for direct human comparison.
\citeA{keshmirian2024biased} directly compared humans and LLMs by asking them to judge the strength of a causal relationship $C \rightarrow B$ as a function of context. Human strength judgments were highest when $C \rightarrow B$ appeared in a chain ($A \rightarrow C \rightarrow B$) versus a fork ($A \leftarrow C \rightarrow B$) or in isolation -- a pattern LLMs matched with a sufficiently high temperature. In contrast, the present work compares human and LLM causal inferences rather than their  strength judgments.

% Contributions
\textbf{Goals and Scope.}
As we increasingly rely on AI-supported decision making, our work aims to contribute to the investigation of biases in causal reasoning and compares those between LLMs and humans using human data previously collected in \citeA
{rehder2017failures}. 
We assess a collider graph where two independent causes influence a shared effect ($C_1 \rightarrow E \leftarrow C_2$). A collider gives rise to four inference types:
predictive inference (see \Cref{fig:comparison_agg_1}), 
unconditional independence  (\Cref{fig:comparison_agg_2}), 
diagnostic inference with both effect present (\Cref{fig:comparison_agg_3}) and absent  
(\Cref{fig:comparison_agg_4}), 
 from which more specific causal reasoning patterns emerge, such as explaining away.  
Using behavioral analyses and modeling with CBNs, we ask if LLMs reason like humans, if they reason normatively, and if their inferences reflect the use of domain knowledge that inheres in their training data.

%%%%%%%%%%%%%%%%%%%%%%%%%%%%%%%%%%%%%%%%%%%%%%%%%%%%%%%%%%%
\section{Methods} \label{sec:methods}
%%%%%%%%%%%%%%%%%%%%%%%%%%%%%%%%%%%%%%%%%%%%%%%%%%%%%%%%%%%
\textbf{Participants.}
We compare the human behavioral data collected in \citeA{rehder2017failures} (Experiment 1,  Model-Only condition, $N = 48$) with judgments gathered from four LLMs --- GPT-3.5 (\colorsquare{gptthreefive}), GPT-4o (\colorsquare{gptfouro}), Claude-3-Opus (\colorsquare{claude}), and Gemini-Pro-1.5 (\colorsquare{gemini}) --- which were prompted with the same inference tasks as humans over their respective APIs.
%The LLMs were tested with five temperature settings $\in \{0.0, .3, .5, .7, 1.0\}$ but we only report results for temperature 0.0 as this ensures consistent and reproducible outputs.
We report results for temperature 0.0 as this ensures consistent and reproducible outputs.

\textbf{Materials.}
The collider causal structure $C_1 \rightarrow E \leftarrow C_2$ was embedded in one of three cover stories from three different knowledge domains (meteorology, economics, and sociology), allowing for a natural language description of the causal structure. 
The three domains were chosen because the undergraduate subjects were expected to be relatively unfamiliar, such that their causal inferences would reflect the causal structure given to them and not idiosyncratic prior knowledge.
% HD: Reviewer expressed confusion about  "the direction of each variable" being counterbalanced; HD rephrased to make it clearer what we mean by counterbalancing
Nevertheless, as an additional safeguard, the adjective describing each variable was counterbalanced  (e.g., in the domain of sociology, some subjects were told that \textit{high} urbanization causes \textit{high} socio-economic mobility, others that it causes \textit{low} socio-economic mobility, etc). In fact, \citeA{rehder2017failures} did not find significant effects of domain or the counterbalancing factor, suggesting that subjects' inferences were not strongly influenced by domain knowledge. An important question we ask here is whether this also holds for the LLMs.
Given a set of observations (a subset of the states of $C_1$, $C_2$, and $E$), both humans and LLMs were asked to provide a likelihood judgment on a continuous scale (0-100) for a specific \textit{query variable} \colordot{inference}.

Below is an \textit{example prompt}  from the sociology domain, matching the visualization in \Cref{fig:example_table} and diagnostic task X  in \Cref{fig:comparison_agg_4}, where the query node (\colordot{inference}) is $C_1=1$  and $C_2$ and the effect $E$ are known to be absent. Note that only the \textit{italicized text} following ``:'' was presented to LLMs in one piece. %The prompt describes a causal mechanism and an observation, followed by an inference task: $ p(C_2 = 1 \mid E = 0, C_1 = 0)$.

\small{ % decrease font size for the example prompt
\begin{itemize}[noitemsep, topsep=0pt]
    \item \textbf{Domain introduction:}
  \textit{Sociologists seek to describe and predict the regular patterns of societal interactions. To do this, they study some important variables or attributes of societies. They also study how these attributes are responsible for producing or causing one another.
        }
     \item \textbf{Variables:} \textit{Here are some variables: Urbanization is the degree to which the members of a society live in urban environments (i.e., cities) versus rural environments. Some societies have high urbanization. Others have normal urbanization.
     Interest in religion is the degree to which the members of a society show a curiosity in religion issues or participate in organized religions. Some societies have low interest in religion. Others have normal interest in religion. 
     Socioeconomic mobility is the degree to which the members of a society are able to improve their social and economic status. Some societies have low socio-economic mobility. Others have normal socio-economic mobility.}
    \item \textbf{Causal mechanism:} \textit{Assume you live in a world that works like this:}
        \begin{itemize}[noitemsep, topsep=0pt]
            \item $C_1=1 \to E=1$: \textit{High urbanization causes high socio-economic mobility.}
            \item $C_2=1 \to E=1$: \textit{Also, low interest in religion causes high socio-economic mobility.}
          
        %\end{itemize}
    \end{itemize}
    \item \textbf{Observation:} \textit{
    %Now suppose you observe the following: 
    Suppose that the society you live in currently exhibits the following:
    normal socio-economic mobility.}
    \item \textbf{Inference task, here $X$} ($p(C_1=1| E=0$)): \textit{Given the observations and the causal mechanism, how likely on a scale from 0 to 100 is high urbanization? 0 means definitely not likely and 100 means definitely likely. Please provide only a numeric response and no additional information.
    % FUll version:
    %Your task is to estimate how likely it is that high interest in religion are present on a scale from 0 to 100, given the observations and causal relationships described. 0 means completely unlikely and 100 means completely certain.  Return your response as raw text in one single line using this exact XML format: <response><likelihood>YOUR_NUMERIC_RESPONSE_HERE</likelihood></response> Replace YOUR_NUMERIC_RESPONSE_HERE with your likelihood estimate between 0 (very unlikely) and 100 (very likely). DO NOT include any other information, explanation, or formatting in your response. DO NOT use Markdown, code blocks, quotation marks, or special characters.
    }
\end{itemize}

\normalsize

To summarize how humans reason with colliders, the empirical findings reported by \citeA{rehder2017failures} are presented in \Cref{fig:main_comparison_agg} (\colorsquare{humans}) alongside the inferences drawn by the LLMs, which are discussed later. The eleven inference tasks (I-XI) are grouped into four types:

\emph{Predictive inferences} in a collider network involve inferring the state of the effect given information about one or more of the causes. Reasoners should judge, for example, that $p(E = 1 \mid C_1 = 0, C_2 = 0) < p(E = 1 \mid C_1 = 0, C_2 = 1) < p(E = 1 \mid C_1 = 1, C_2 = 1)$.
\Cref{fig:comparison_agg_1} reveals that human reasoners in fact exhibit this pattern, indicated by a monotonically increasing slope, confirming that they made use of the causal knowledge on which they were instructed.

\emph{Independence of causes} is another property of colliders. Because in CBNs exogenous causes are stipulated to be uncorrelated, reasoners should judge that the presence of one cause should not affect the likelihood of the other: $p(C_1 = 1 \mid C_2 = 1) = p(C_1 = 1 \mid C_2 = 0)$, which would be reflected as a flat line in \Cref{fig:comparison_agg_2}. Instead, humans judged  that $p(C_1=1|C_2=1) > p(C_1=1|C_2=0)$. This is an instance of the well-known \textit{Markov violations} that characterize how humans reason with numerous causal network topologies involving generative relations \cite{davis2020mutation}. 
Markov violations have been characterized as an \textit{associative bias}
%Rehder \& Waldmann, 2017,
 \cite<or what>[referred to as a \textit{rich-get-richer} bias]{rehder2017failures}, where the presence of one causal variable makes another supposedly independent variable more likely. 
%Both weak explaining away and 
% mentioned in next paragraph
Markov violations with collider graphs have been documented in multiple studies \cite<see>[for a review]{davis2020mutation}.

\textit{Diagnostic inferences} involve inferring the state of one cause given the effect and possibly the other cause. In collider structures with independent causes and the effect present, this gives rise to \textit{explaining away}, where observing that one cause is present/absent should lower/raise the probability of the other cause. This phenomenon stipulates two conditions. (i) Explaining away proper is when observing one cause reduces the likelihood of the other, e.g., $p(C_1 = 1 \mid E = 1, C_2 = 1) < p(C_1 = 1 \mid E = 1)$. (ii) \textit{Augmentation} arises when observing the absence of a cause increases the likelihood of the other, e.g., $p(C_1 = 1 \mid E = 1, C_2 = 0) > p(C_1 = 1 \mid E = 1)$. 
%In collider structures with independent causes, this gives rise to \textit{explaining away}, where observing one cause reduces the likelihood of the other, conditional on the effect: $p(C_1 = 1 \mid E = 1, C_2 = 1) < p(C_1 = 1 \mid E = 1) < p(C_1 = 1 \mid E = 1, C_2 = 0)$. We refer to this pattern as involving two conditions: (i) the likelihood of $C_1 = 1$ increases when $C_2$ is known to be absent $p(C_1 = 1 \mid E = 1) < p(C_1 = 1 \mid E = 1, C_2 = 0)$, and (ii) it decreases when $C_2$ is present ($p(C_1 = 1 \mid E = 1, C_2 = 1) < p(C_1 = 1 \mid E = 1) $). 
\Cref{fig:comparison_agg_3} demonstrates that humans exhibited the overall explaining away pattern, consistent with the expected monotonically increasing slope under conditions (i) and (ii). However, the effect is weak (i.e., the slope is shallow), aligning with theoretical work showing that explaining away is often attenuated relative to normative expectations \cite{davis2020mutation, rehder2024inhibitory}. 
%Notably, when effect $E$ is absent (\Cref{fig:comparison_agg_4}), the explaining away pattern disappears entirely for humans.
% TODO: HD: describe E=0 diag. reasonning, DONE
If the causal relations in the experiment are assumed to be deterministically sufficient and necessary, then the absence of the effect should imply a zero probability for the presence of its causes. Yet \Cref{fig:comparison_agg_4} revealed human likelihood judgments for $C_1=1$  well above zero, suggesting they did not fully endorse this deterministic framing.

\textbf{Procedure.} 
%A key contribution of this work is the creation of a causal inference task dataset, enabling direct comparisons between human causal inference judgments  collected in \citeA{rehder2017failures} and LLMs.The dataset is designed to closely replicate the experimental conditions of \citeA{rehder2017failures} (Experiment 1, Model-Only condition) with some  notable differences: The procedure for humans consisted of two phases. In the learning phase subjects were presented and tested on the domain knowledge, including the causal mechanisms. In the testing phase they were presented with each of the inference tasks in random order. A graphical representation of the collider structure remained on the screen during testing. In contrast, for the LLMs each textual prompt included all the domain knowledge and a single inference task. Whereas humans provided their probability judgments using a slider ranging from 0 to 100 with default setting=50.0, LLMs were instructed to provide a numerical answer $\in \{0.0, 100.0\}$. 
A key contribution of this work is the creation of a causal inference task dataset enabling direct comparisons between human causal inference judgments collected in \citeA{rehder2017failures} and LLMs.
The dataset closely replicates the experimental conditions of \citeA{rehder2017failures} (Experiment 1, Model-Only condition) with some notable differences:
The human procedure consisted of two phases. In the learning phase, subjects were presented and tested on domain knowledge, including causal mechanisms. In the testing phase, they completed each inference task in random order. A graphical representation of the collider structure remained visible during testing.
In contrast, each LLM prompt included all domain knowledge and a single inference task.
Whereas humans provided probability judgments using a 0–100 slider (default = 50.0), LLMs were instructed to provide a numerical answer $\in {0.0, 100.0}$.
%%%% Note: include repeated 5 times ... when talking about other temperature settings than 0.0. since the same script was used for all temperature settings, also for temp=0.0 the query was repeated 5 times. Note that data per each of the 528 unitque experimental conditions (LLMxtaskxDomainxCounterbalance condition where temp=0.0) reveals about only have have the responses following the expected zero variation, partially confirming that temp=0.0 is indeed yielding deterministic ouputs. 235 of the above conditions had a std \neq 0.
%%%% end note

%%% Model fitting:
% Do LLMs reason consistently?
% include tex file
%\input{maybe_someday_useful_text/model_fit_text_hanna/model_fit_methods.tex}
  
%%%%%%%%%%%%%%%%%%%%%%%%%%%%%%%%%%%%%%%%%%%%%%%%%%%%%%%%%%%
%%%%%%%%%%%% RESULTS %%%%%%%%%%%%%%%%%%%%%%%%%%
\section{Results}   %%%%%%%%%%%%%%%%%%%%%%%%%%
\begin{figure*}[htbp]
    \centering
    \begin{subfigure}[t]{0.1\textwidth}
        \centering
        \includegraphics[width=\textwidth]{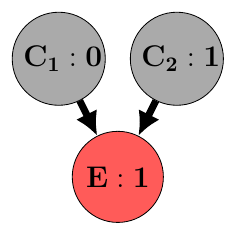}
        \caption{Reference Graph: task II}
 
        \label{fig:graph}
    \end{subfigure}
    \hfill
    \begin{subfigure}[t]{0.22\textwidth}
        \centering
        \includegraphics[width=\textwidth]{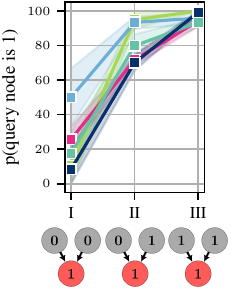}
        \caption{Predictive Inference}
 
        \label{fig:comparison_agg_1}
    \end{subfigure}
    \hfill
    \begin{subfigure}[t]{0.22\textwidth}
        \centering
        \includegraphics[width=\textwidth]{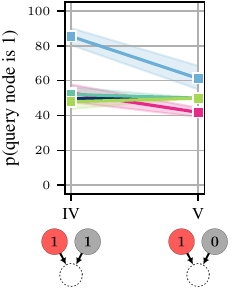}
        \caption{Independence of $C_1, C_2$}
        \label{fig:comparison_agg_2}
    \end{subfigure}
    \hfill
    \begin{subfigure}[t]{0.22\textwidth}
        \centering
        \includegraphics[width=\textwidth]{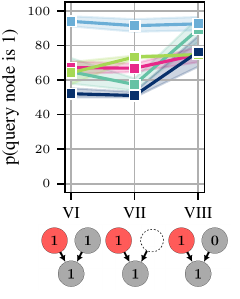}
        \caption{Diagnostic Inference \\ \hspace*{11mm} $(E=1)$}
        \label{fig:comparison_agg_3}
    \end{subfigure}
    \hfill
    \begin{subfigure}[t]{0.22\textwidth}
        \centering
        \includegraphics[width=\textwidth]{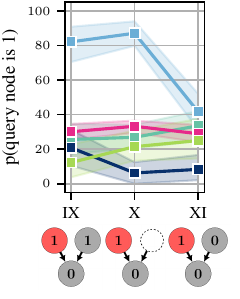}
        \centering
        \caption{Diagnostic Inference \\\hspace*{11mm}  $(E=0)$}
        \label{fig:comparison_agg_4}
    \end{subfigure}
    \caption{\textbf{Aggregated across all domains:} Likelihood judgments that query node \colordot{inference} has value 1  $\in \{0,100\}$ with bootstrapped 95\% confidence intervals of humans \colorsquare{humans} and LLMs (GPT-3.5 \colorsquare{gptthreefive}, GPT-4o \colorsquare{gptfouro}, Claude \colorsquare{claude}, and Gemini \colorsquare{gemini})  for each inference task (I-XI), aggregated across counterbalancing conditions and domains for temperature value 0.0 (most deterministic). Graphs on the x-axis visualize the conditional probability of the inference tasks (I-XI) where the  nodes are colored according to:  \colordot{inference} $\to$ query node that the question is asked about; \colordot{observed} $\to$ observed $\in \{0,1\}$; and \dashedcircle{}   $\to$  no information.} 
    \label{fig:main_comparison_agg}
\end{figure*}

% r_s updated 05/06
\textbf{Comparison of LLMs and Humans.} As an initial assessment of LLM-human reasoning alignment, we computed the Spearman correlation between their inferences and those of humans in each domain. \Cref{tab:correlations} reveals correlations that are positive and substantial in magnitude, indicating the LLMs are exhibiting a degree of human-like performance on the causal reasoning tasks. The highest average correlations were displayed by Gemini \colorsquare{gemini} ($r_s = .763$), followed by Claude \colorsquare{claude} ($r_s = .677$) and GPT-4o \colorsquare{gptfouro} ($.658$). Least aligned was GPT-3.5 \colorsquare{gptthreefive} ($r_s=.390$). This pattern was observed in all three domains.

% r_s updated 05/04
\begin{table}[h]
    \centering
        \caption{Spearman correlations $r_s$ between human and LLM inferences in each domain / across domains (pooled).}
        \begin{small}
    \resizebox{\columnwidth}{!}{ 
    \begin{tabular}{lcccc}
        \toprule
         & \multicolumn{3}{c}{Domain} &  \\
        \cmidrule(lr){2-4}
        Model & Economy ($r_s$) & Sociology ($r_s$) & Weather ($r_s$) & Pooled \\
        \midrule
        Claude \colorsquare{claude}  & .641 & .739 & .755 & .677 \\
        GPT-4o \colorsquare{gptfouro}   & .618 & .506 & .767 & .658 \\
        GPT-3.5 \colorsquare{gptthreefive}  & .390 & .473 & .313 & .390 \\
        Gemini \colorsquare{gemini}  & \textbf{.713} & \textbf{.743} & \textbf{.855} & \textbf{.763} \\
        \bottomrule
    \end{tabular}
    }\end{small}
    \label{tab:correlations}
\end{table}

%%%% Line plot description

\Cref{fig:main_comparison_agg} presents the LLMs' responses to the four inference tasks averaged over conditions. 
The main finding is that all LLMs except GPT-3.5 \colorsquare{gptthreefive}  provided sensible judgments for all inference tasks. 
Each task reveals distinct reasoning patterns across agents. 

 %While we tested outputs across a range of temperatures (0.0 - 1.0), model sensitivities varied—for instance, GPT-3.5 was invariant, while Gemini-pro had 32 significant variations. By limiting the analysis to temperature=0.0, we ensure a consistent and interpretable comparison, avoiding confounds from stochastic variability at higher temperatures.
 %The variability in LLM responses stems from the counterbalance conditions, which were aggregated over  in the analysis.

% - LLMs can do the task
\emph{Predictive inferences} (\cref{fig:comparison_agg_1}, I-III) for the LLMs were a monotonic increasing function of the number of causes present, similar to the human judgments. This indicates that the LLMs were sensitive to the most rudimentary aspect of the task, namely, that causes make their effects more likely. Predictive inference is the only inference type where GPT-3.5 \colorsquare{gptthreefive} provided sensible judgments. 

\emph{Independence of causes} (\cref{fig:comparison_agg_2}), IV-V) means that the state of one cause should not affect the likelihood of the other (i.e., a flat line). 
%Thus, $C_1$ should be unaffected by whether $C_2$ was present (i.e., a flat line). %CW: Could we say something here about what pattern to expect (normatively)? HD: Done above, by stating that we would expect a flat line
 GPT-3.5 \colorsquare{gptthreefive} violated this principle by judging $p(C_1=1|C_2=1) > p(C_1=1|C_2=0)$ even more egregiously than humans. 
Conversely, Claude \colorsquare{claude}, GPT-4o \colorsquare{gptfouro}, and Gemini \colorsquare{gemini} reasoned  normatively, by respecting the independence of causes, indicated by a flat line. 

% TODO: HD: merge with expl. away description further down. Done!
\emph{Effect-Present Diagnostic Inference} (\cref{fig:comparison_agg_3}, VI–VIII) assessed explaining away via the slope of inferred probabilities reflected by a positive slope between tasks VI and VII if (i) holds. Gemini-Pro  \colorsquare{gemini} showed the strongest effect, followed by both humans \colorsquare{humans} and GPT-4o \colorsquare{gptfouro} with weak (i).   Claude  \colorsquare{claude} and  GPT-3.5 \colorsquare{gptthreefive} violated  explaining away,  indicated by negative slope for (i). 
Conversely, GPT-4o and Claude showed strong augmentation (ii), assigning higher likelihood to $C_1=1$ when the alternative cause was absent, indicated by a positive slope between tasks VII and VIII. Gemini and GPT-3.5 exhibited numerically weak augmentation ($<2$ points), and no model fully satisfied both conditions.

% TODO: HD: merge with effect-absent description further down
\emph{Effect-Absent Diagnostic Inference} (\cref{fig:comparison_agg_4}, IX-XI)
has all agents produce lower ratings for the cause, with GPT-4o \colorsquare{gptfouro} and Claude \colorsquare{claude} producing the lowest ratings across all conditions and Gemini \colorsquare{gemini} seeming to be closest aligned with humans \colorsquare{humans}. 
While humans and Gemini \colorsquare{gemini} are more likely to assign ratings in the middle of the scale, 
GPT-4o \colorsquare{gptfouro} is most inclined to assign a rating of 0 
and treated the causal relations as closer to necessary and sufficient than any other agent. This interpretation is supported by the model fitting that follows, which yielded especially large estimates of the strengths of the causal relations for GPT-4o \colorsquare{gptfouro} (see \Cref{fig:fitted_parameters}).

% I've updated the numbers and text. But the LLM/human differences are not as dramatic as they were, so this could easily be deleted.
Note that the responses of three of the four LLMs in \Cref{fig:main_comparison_agg} exhibited a greater range than the humans. The difference between the highest and lowest judgment was 95.0, 91.7, and 75.8 for GPT-4o, Gemini, and Claude, respectively, as compared to 66.0 for the humans. This tendency might stem from the experimental setup. Whereas LLMs were prompted to generate a single numeric value, humans responded using an interactive slider that defaulted to 50. This default could have introduced a motor bias that encouraged responses near the middle of the scale. The responses of GPT-3.5 exhibited the narrowest range (54.2).
% summary
These inference patterns suggest LLMs capture core causal reasoning principles and are aligned with human responses to a considerable degree. 
% TODO: confirm if this still holds for new model fits
Some LLMs' reasoning patterns in \Cref{fig:main_comparison_agg} reveal that causal relations were treated as close to necessary and sufficient (e.g., GPT-4o \colorsquare{gptfouro}), which is also supported later when we fit CBNs (see \Cref{fig:fitted_parameters}).

\textbf{CBN Model Fitting.}
Next, we evaluate LLMs and humans against normative inferences from a causal Bayes net (CBN). Since agents received only verbal descriptions, the CBN's parameters $\theta_M$ were treated as free parameters and fit to the data. These parameters were the causes' prior probabilities $w_C$, representing $p(C_1)$ and $p(C_2)$, the causal strength parameters $w_{C_1,E}$ and $ w_{C_2,E}$, representing the strength of $C_{1} \to E$ and $C_{2} \to E$, and $w_E$, representing the influence of any exogenous causal influence on $E$.

The CBN is used to derive a joint probability distribution which was then used to derive the conditional probability appropriate for that task. For a collider causal graph $C_1 \rightarrow E \leftarrow C_2$, the joint distribution was derived assuming that $p(C_1,C_2,E) = p(E|C_1,C_2)p(C_1)p(C_2)$ and that $p(E=1|C_1,C_2) = 1/(1 + \exp(-(C_1w_{C_1,E} + C_2w_{C_2,E} + w_E)))$, where $C_1$ and $C_2$ are each coded as 1 when present and $-1$ when absent.\footnote{Note in this literature it is common to assume ``noisy logical'' generating functions, such as the noisy-OR function introduced in the 
PowerPC theory  of causal learning by \citeA{cheng1997covariation}. We report fits using the logistic generating function as it consistently yielded better fits to these data sets.} The CBNs were fit to each agent's set of causal judgments by identifying parameters that minimized squared error. Fits were carried out via an initial grid search followed by optimization.

We fit two variants of the basic collider CBN. The first assumed that the two causal strengths were equal, that is, $w_{C_1,E} = w_{C_2,E} = w_{C,E}$. Thus, the parameters of this model were $w_C$, $w_{C,E}$, and $w_E$. A 4-parameter variant allowed the strength of the causal relations to differ by fitting $w_{C_1,E}$ and $w_{C_2,E}$ separately. $w_C$ was constrained to the range [0, 1] and the causal strength parameters were constrained to [$-3$, 3]. For the human data, these CBNs were fit to each subject. For the LLMs, they were separately fit to the judgments in each of the 3 domains $\times$ 4 counterbalancing $= 12$ conditions. 

% TODO update model fit number, DONE!
 \Cref{tab:ModelFits-3v4Params} presents the CBNs' best fitting parameters averaged over conditions for each agent. Several trends emerge. The correlations between the observed judgments and those predicted by the fitted CBNs were substantial for all the LLMs, ranging from $0.503$ to $0.879$. Notably, for Gemini-Pro, GPT-4o, and Claude-3 those correlations were all greater than $0.82$ and so greater than those observed for the humans ($0.77$). They also exhibited more favorable model losses, defined as the average absolute prediction error on the $0$–$100$ scale,  than the humans. That is, if CBNs are accepted as the normative standard, these LLMs exhibited more accurate causal reasoning than the humans. In contrast, GPT-3.5 performed worse than both humans and other LLMs, with correlations below $0.560$ and model losses above $14$.

\begin{table}[t!]
\setlength{\tabcolsep}{4pt}
\begin{tiny}
\centering
\caption{Fits of causal Bayes nets (CBN) by agent.}
\begin{tabular}{lllllllllll}
\toprule 
    & &
    \multicolumn{5}{c}{Average Parameter Estimates} & 
    \multicolumn{3}{c}{Measures of Fit}\\
    Agent & $NP$ & $w_C$ & $w_{C,E}$ & $w_{C_1,E}$ & $w_{C_2,E}$ & $w_{E}$ & $R$ & $AIC$ & Loss \\
\midrule
   Humans \colorsquare{humans}
   & \textbf{3} & \textbf{.528} & \textbf{1.06} & & & \textbf{0.91} &  \textbf{.770}  & \textbf{114.4}  & \textbf{11.8} \\
   & 4 & .529 & & 1.09 & 1.04 & 0.92 & .783 & 115.4 & 11.5\\
       & & & & & & & & & \\
   Gemini-Pro-1.5  \colorsquare{gemini}
   & \textbf{3} & \textbf{.553} & \textbf{1.55} & & & \textbf{1.87} & \textbf{.877} & \textbf{109.6} & \textbf{10.2} \\
   & 4 & .556 & & 1.57 & 1.59 & 1.96 & .877 & 110.5 &  10.2\\
       & & & & & & & & & \\
   GPT-3.5   \colorsquare{gptthreefive}
   & 3 & .843 & 0.60 & & & 1.76 & .503 & 123.4 & 15.4 \\
   & \textbf{4} & \textbf{.845} & & \textbf{0.61} & \textbf{0.59} & \textbf{1.83} & \textbf{.558} & \textbf{123.3} &  \textbf{14.6}\\
       & & & & & & & & & \\
   GPT-4o \colorsquare{gptfouro}
   & \textbf{3} & \textbf{.438} & \textbf{1.66} & & & \textbf{1.31 }& \textbf{.879} & \textbf{107.1} & \textbf{9.88} \\
   & 4 & .436 & & 1.74 & 1.53 & 1.32 & .881 & 107.6 & 9.88\\
       & & & & & & & & & \\
   Claude-3-Opus \colorsquare{claude}
   & \textbf{3} & \textbf{.555 }& \textbf{1.26} & & & \textbf{1.16} & \textbf{.829} & \textbf{110.6} & \textbf{11.2} \\
   & 4 & .554 & & 1.32 & 1.21 & 1.17 & .839 & 111.7 & 10.8\\
\bottomrule
\end{tabular}%}
\label{tab:ModelFits-3v4Params}
\centering
\caption*{\footnotesize\emph{Note}: $NP$ = Number of model parameters. $AIC$ = Akaike Information Criterion, used to choose winning CBN in bold.}
\end{tiny}
\vspace{-2em}
\end{table}

Regarding the contrast between the 3- and 4-parameter CBNs, the human data did not benefit from the extra causal strength parameter. This result is consistent with past analyses of these data showing that neither domain nor the counterbalancing factor had a significant effect on subjects' judgments \cite{rehder2017failures}. Turning to the LLMs, only GPT-3.5 yielded a better fit with two causal strength parameters. Although we expected that the LLMs might be more likely to assume causal relations of different strength by using the knowledge they have about economics, meteorology, and sociology, the fitted parameter values in \Cref{tab:ModelFits-3v4Params} indicate that they were no more likely to do so than the humans.\footnote{We also fit CBNs in which the two causes $C_1$ and $C_2$ each had their own parameter representing their prior probability. Generally, these models did not yield a better fit than the models with a single $w_C$ parameter. The one exception was GPT-3.5, but as this model yielded relatively poor fits, we do not discuss this result further.} A detailed investigation of the effect of domain on the parameter estimates would offer further insight into agents’ sensitivity to contextual and linguistic variation in the causal cover stories but is beyond the scope of the current work.

  %To  provide a more granular perspective on agent behavior than the averaged results in \Cref{tab:ModelFits-3v4Params}, \Cref{fig:fitted_parameters} visualizes the fitted parameter distributions from the 4-parameter CBN. Each violin plot summarizes the distribution of agent-specific  parameter values across domains and counterbalancing conditions. The figure includes the absolute difference between the two causal strength parameters, $|w_{C_1,E} - w_{C_2,E}|$. Parameter $w_C$, which  represents the prior over the causes (i.e., $p(C_1)$ and $p(C_2)$), was clustered around  $0.5$ with means across agents ranging from 0.43 (GPT-4o) to 0.85 (GPT-3.5) and was the parameter exhibiting the least variation for all agents. The causal strength parameters $w_{C_1,E}$ and $w_{C_2,E}$ showed greater variability. GPT-4o was most often best fit by high values, with median estimates of $1.59$ and $1.30$, suggesting strong deterministic assumptions. Although Claude and GPT-4o had a higher $w_{C_1,E}$, $w_{C_2,E}$  parameter range than Gemini ($0.206$–$2.70$), Gemini had the highest median ($1.65$). Humans were the only group occasionally best fit by negative values for $w_{C_2,E}$ (range: $-0.57$ to $3.00$), indicating possible inhibitory interpretations. GPT-3.5 was most often best fit by small causal strength parameters didn't show as much variation as any of the other agents.  

To provide a more granular view of agent behavior than the averaged results in \Cref{tab:ModelFits-3v4Params}, \Cref{fig:fitted_parameters} shows fitted parameter distributions from the 4-parameter CBN. Each violin plot summarizes agent-specific parameter values across domains and counterbalancing. The figure also includes the absolute difference between the two causal strength parameters, $|w_{C_1,E} - w_{C_2,E}|$. Parameter $w_C$, representing the prior over causes $p(C_1)$ and $p(C_2)$, clustered around $0.5$, with agent means ranging from 0.43 (GPT-4o) to 0.85 (GPT-3.5), and showed the least variation across agents. The causal strength parameters $w_{C_1,E}$ and $w_{C_2,E}$ showed greater variability. GPT-4o was most often best fit by high values, with median estimates of $1.59$ and $1.30$, suggesting strong deterministic assumptions. Although Claude and GPT-4o had a broader $w_{C_1,E}$, $w_{C_2,E}$ range than Gemini ($0.206$–$2.70$), Gemini had the highest median ($1.65$). Humans were the only group occasionally best fit by negative values for $w_{C_2,E}$ (range: $-0.57$ to $3.00$), suggesting possible inhibitory interpretations. GPT-3.5 showed the least variability of any agent and favored smaller values (range: $.05$ to $1.73$, median: $.62$ and  $.43$).
The absolute difference between causal strengths $|w_{C_1,E} - w_{C_2,E}|$ was generally small, ranging from $0.0$ to $1.03$, suggesting that, within a domain, $C_{1} \to E$ and $C_{2} \to E$ were treated as about equally strong. Like the causal strengths, parameter $w_E$, reflecting sensitivity to exogenous causes, also exhibited substantial variability over domains (range: $-.39$ to $3.0$).

\begin{figure}[h]
    \includegraphics[width=\columnwidth]{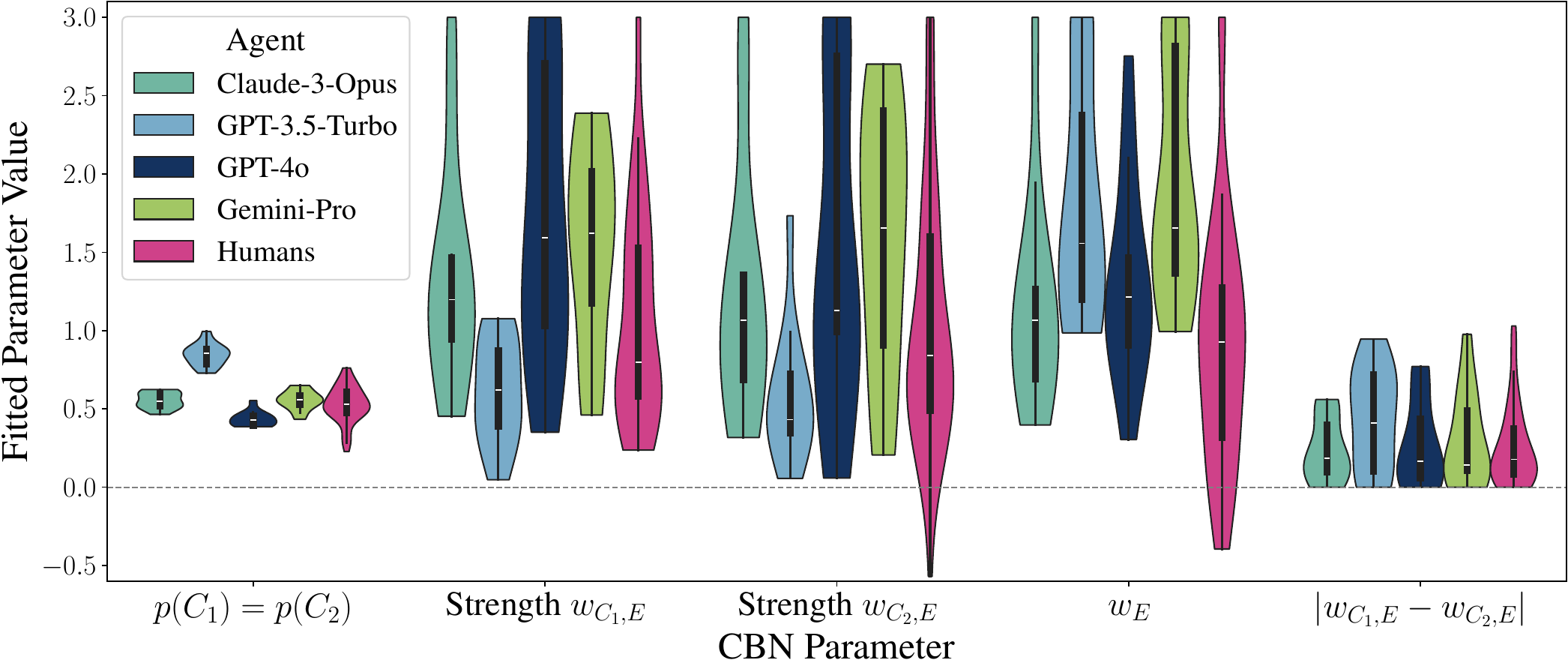}
    \centering
    %\caption{Fitted parameter distributions for each agent under the 4-parameter CBN model. Violin plots reflect aggregated fits across domains and counterbalancing conditions. The white bar denotes the median, and the box spans the 25th to 75th percentiles. The rightmost plot quantifies asymmetry in inferred causal strengths via the absolute difference $|w_{C_1,E} - w_{C_2,E}|$.}
    % shorter version:
    \caption{Fitted parameter distributions for each agent under the 4-parameter CBN model. Violin plots reflect aggregated fits across domains and counterbalancing. The white bar denotes the median; the box spans the 25th to 75th percentiles. The rightmost plot quantifies asymmetry in inferred causal strengths via the absolute difference $|w_{C_1,E} - w_{C_2,E}|$.}
    \label{fig:fitted_parameters}
\end{figure}

\textbf{Fitting a Psychological Model.} 
We also fit the LLM inferences with a model proposed as an account human causal reasoning, the \textit{mutation sampler} \cite{davis2020mutation}. The mutation sampler is an example of a \textit{rational process model}, an algorithm that yields normative responses when cognitive resources are unlimited but that produces errors when they are not \cite{johnson2016computational, lieder2012burn, vul2014one}. The mutation sampler carries out MCMC sampling over a causal graph's \textit{state space} and draws inferences on the basis of samples. But because sampling begins at one of the graph's \textit{prototypes states} (when causal relations are all generative, the states where variables are all present or all absent), errors are introduced when the number of samples drawn is limited. \citeA{davis2020mutation} showed that the associative bias induced by the prototypes allowed the mutation sampler to account for the independence violations that arise in a wide variety of network topologies and the weak explaining away that arises when reasoning about collider graphs. 

The mutation sampler was fit to the human data and the four LLMs.  \Cref{tab:ModelFits-MutSamp} presents the improvement for each data set relative to the 3-parameter CBN in  \Cref{tab:ModelFits-3v4Params} realized by adding the mutation sampler's \textit{chain length} free parameter $\lambda$ representing the number of MCMC samples. Replicating past findings, the mutation sampler yielded a better fit to the human data (according to $AIC$) compared to the 3-parameter CBN \cite{davis2020mutation}. 
In contrast, \Cref{tab:ModelFits-3v4Params} shows that it generally did \textit{not} yield a better fit for the LLMs (GPT-3.5 was the only exception). Apparently, the LLMs were less susceptible to the associative reasoning processes that influence people's causal inferences, a conclusion supported by the fitted chain length parameters $\lambda$ shown in \Cref{tab:ModelFits-MutSamp}. Because the impact of the starting point diminishes as the chain length grows, that the LLM fits exhibited relatively large chain lengths indicates that the associative influence induced by the prototypes had little impact on the LLMs' judgments. 
%CW: could this be rephrased to be more intuitive? BR: I tried.

\begin{comment}
\begin{table}[ht]
\caption{Mutation sampler fits.}
\label{ModelFits-MutSampler}
\centering
\begin{small}
\begin{tabular}{lccccc}
\toprule 
    Agent & $NP$ & $\lambda$ & $R$ & $AIC$ & $Loss$\\
\midrule
   Human & \textbf{4} & \textbf{3.7} & \textbf{.810} & \textbf{113.0} & \textbf{10.8} \\
   Gemini-Pro & 4 & 38.6 & .881 & 109.7 & 10.0 \\
   GPT-3.5-Turbo & \textbf{4} & \textbf{37.9} & \textbf{.776} & \textbf{113.1} & \textbf{16.4} \\
   GPT-4o & 4 & 26.0 & .881 & 108.4 & 9.81 \\
   Claude-3 & 4 & 43.9 & .826 & 114.1 & 11.5 \\
\bottomrule 
\end{tabular}
\end{small}
\end{table}
\vspace{-.2em}
\end{comment}

\begin{table}[t!]
\setlength{\tabcolsep}{5pt}
\begin{tiny}
\centering
\caption{Fits of the 4-parameter mutation sampler.}
\begin{tabular}{llllllllll}
\toprule 
    & &
    \multicolumn{4}{c}{Average Parameter Estimates} & 
    \multicolumn{3}{c}{Measures of Fit}\\
    Agent & $NP$ & $w_C$ & $w_{C,E}$ & $w_{E}$ & $\lambda$ & $R$ & $AIC$ & Loss \\
\midrule
   Humans \colorsquare{humans}
   & 3 & .528 & 1.06 & 0.91 & & .770  & 114.4  & 11.8 \\
   & \textbf{4} & \textbf{.523} & \textbf{0.94} & \textbf{0.83} & \textbf{3.7} & \textbf{.810} & \textbf{113.0} & \textbf{10.8} \\
   \\
   Gemini-Pro-1.5  \colorsquare{gemini}
   & \textbf{3} & \textbf{.553} & \textbf{1.55} & \textbf{1.87} & & \textbf{.877} & \textbf{109.6} & \textbf{10.2} \\
   & 4 & .576 & 1.43 & 1.97 & 38.6 & .881 & 109.7& 10.0 \\
   \\
   GPT-3.5   \colorsquare{gptthreefive}
   & 3 & .843 & 0.60 & 1.76 & & .503 & 123.4 & 15.4 \\
   & \textbf{4} & .897 & \textbf{-0.24}& \textbf{2.46} & \textbf{37.9} & \textbf{.776} & \textbf{113.1} & \textbf{16.4} \\
   \\
   GPT-4o \colorsquare{gptfouro}
   & \textbf{3} & \textbf{.438} & \textbf{1.66} & \textbf{1.31 }& & \textbf{.879} & \textbf{107.1} & \textbf{9.88} \\
   & 4 & .398 & 1.50 & 1.09 & 26.0 & .881 & 108.4 & 9.81 \\
   \\
   Claude-3-Opus \colorsquare{claude}
   & \textbf{3} & \textbf{.555 }& \textbf{1.26} & \textbf{1.16} & & \textbf{.829} & \textbf{110.6} & \textbf{11.2} \\
   & 4 & .569 & 1.21 & 1.20 & 43.9 & .826 & 114.1& 11.5\\
\bottomrule
\end{tabular}%}
\label{tab:ModelFits-MutSamp}
\centering
\caption*{\footnotesize\emph{Note}: The 3-parameter CBN is included for comparison.}
\end{tiny}
\vspace{-2em}
\end{table}

%%%%%%%%%%%%%%%%%%%%%%%%%%%%%%%%%%%%%
\section{Discussion}
%%%%%%%%%%%%%%%%%%%%%%%%%%%%%%%%%%%%% 
We compared the causal reasoning abilities of large language models to those of people. In \citeA{rehder2017failures} undergraduates were taught hypothetical causal knowledge consisting of three variables that formed a collider causal graph and then were asked to draw simple causal inferences. 
%TODO: HD: mention limitations between human and llm prompts; Done!
Our first main finding is that given  the same information\footnote{Humans underwent a learning phase with extensive exposure to background information, whereas the inference task was limited to a single screen displaying only essential details. In contrast, LLMs were presented with both the learning and testing phases simultaneously in one long prompt. What constitutes an equivalent input for LLMs remains an open question.},  most LLMs tested can do the task. 
% predicitve inference results
That is, after being told that the presence of one variable $C$ causes the presence of another $E$, LLMs will judge the effect $E$ is more likely when a cause  $C$ is present versus absent (and vice versa). 
% updated with new values
Indeed, across all domains and tasks, the Spearman correlation $r_s$ between LLM  and human inferences ranged from $.313$ to $.855$.

 %  independence of causes
 % HD: updated reflecting new results
%A collider structure also entails that the two causes should be independent. As discussed, human reasoners often violate independence, which in a collider with generative causes means that the two causes are treated as positively correlated \cite{davis2020mutation}.
%In other words, human causal inferences also exhibit a degree of ``associative thinking." GPT-3.5's also treated the causes as positively correlated, consistent with the interpretation of their explaining away inferences offered above. 
Collider structures imply that causes are independent, yet human judgments  often violate independence, reflected as a perceived positive correlation, consistent with associative reasoning \cite{davis2020mutation}. \Cref{fig:comparison_agg_2} shows that GPT-3.5 exhibited a similar but stronger violation. By contrast, Claude, Gemini, and GPT-4o adhered  closely to the independence assumption, assigning uniform likelihoods ($\approx50$) regardless of the status of the alternative cause. 
% Commenting out below to save space
%GPT-3.5 further diverged by assigning markedly higher likelihoods (60–90), in contrast to all other models and human participants.

% TODO: HD: elaborate on partial explaining away and discuss conditions that need to be met, in conjunction with earlier 
% HD: Done!
% HD: shorten and move up to results
  %The LLMs diverged in how they satisfied the two conditions that define explaining away (\Cref{fig:comparison_agg_3}). Only Gemini exhibited explaining away proper (i) by assigning a lower likelihood to a cause when the alternative cause was present. Only GPT-4o and Claude exhibited strong augmentation (ii) by assigning a higher likelihood to the cause when the alternative was absent. (Gemini and GPT-3.5 also exhibited an effect of augmentation but one that was numerically very weak, i.e., less than 2 points on the 0-100 scale). That is, none of the LLMs exhibited fully satisfactory explaining away. Thus, whereas the LLMs generally exhibited impressive correlations with humans and their fitted CBNs, they did not capture some of the more subtle reasoning patterns implied by a collider graph. 

% Commenting out below to save space
  %An important property of collider structures is that they entail \textit{explaining away}.
  The LLMs varied in how they exhibited explaining away, with no model fully capturing both defining conditions (i) and (ii) (\cref{fig:comparison_agg_3}).
  Gemini-Pro was the only model to show a clear instance of condition (i), reducing the likelihood of one cause when the alternative was present compared to when there was no information \dashedcircle{} about the alternative cause (\cref{fig:comparison_agg_3}, VI, VII). 
  Conversely, GPT-4o and Claude demonstrated strong augmentation (condition (ii)), increasing the likelihood of one cause when the other was absent (\cref{fig:comparison_agg_3},  VII, VIII). 
  %That is, none of the LLMs exhibited fully satisfactory explaining away. 
  Thus, whereas the LLMs generally exhibited impressive correlations with humans and their fitted CBNs, they did not capture some of the more subtle reasoning patterns implied by a collider graph. 
  
  %Gemini uniquely satisfied condition (ii), assigning lower likelihood to $C_1 = 1$ when the alternative cause was present, but failed to meet condition (i), assigning equal likelihood regardless of whether the alternative cause was known to be absent or whether there was no information about. In contrast, GPT-4o and Claude satisfied condition (i): both increased likelihood when the alternative cause was known to be absent. However, neither satisfied condition (ii): GPT-4o assigned equal likelihood whether the alternative cause was present or not, while Claude reversed the expected pattern by assigning higher likelihood to $C_1 = 1$ when the alternative cause was present -- directly contradicting condition (ii).
 %One interpretation of this result is that the two latter models were reasoning more ``associatively,'' that is, without regard to causal semantics of a collider network. At least on this one test of ``causal understanding,''  GPT-3.5 failed.

% Effect absent reasoning: updated with new data
A collider structure also supports diagnostic reasoning in the absence of the effect (see \Cref{fig:comparison_agg_4}). Note that, if the causal relations are interpreted as deterministically sufficient (the cause always produces the effect), then the likelihood of the causes should be zero when the effect is absent. GPT-3.5 deviated sharply from this prediction, providing judgments of between 40 and 90 (and consistent with its relatively weak fitted causal strength parameters in \Cref{tab:ModelFits-3v4Params}). In contrast, the other LLMs provided lower judgments for these inferences (consistent with their larger causal strength parameters). Of all the LLMs, the responses of GPT-4o were most consistent with deterministically sufficient causal relations.

%Normatively, if the effect is absent, the likelihood of either cause should be zero. GPT-3.5 deviated sharply from this expectation, assigning likelihoods between 40 and 90. The other models provided lower estimates, with GPT-4o most closely approximating zero. Claude's responses aligned more closely with human judgments, falling between 20 and 40.

% model fits
In addition to humans, we compared LLMs to the normative inferences of fitted CBNs. Correlations between LLM and normative inferences ranged from .503 to .881, versus $\approx$ .77 for humans. GPT-3.5 showed the weakest correlations -- lower than those of humans --  whereas GPT-4o, Gemini, and Claude showed the highest, \textit{exceeding} human correlations. Computational model fitting revealed that one reason for the better performance of the latter models is that they didn't exhibit the associative bias that plagues human casual reasoning. 
%this is obsolete
%The parameters derived from the CBN fits also provided insight into how the agents were reasoning in the different domains. As mentioned, the materials were drawn from domains about which undergraduates were expected to know little. Consistent with this expectation, the standard deviations associated with the human's fitted CBN parameters were relatively low (e.g., for the average human reasoner the difference between estimated strength of the two causal relations differed by only .12). In contrast, the standard deviations for with the LLM's fitted model parameters were much higher (e.g., the difference between the two causal strengths was at least .45). These results provide evidence that the LLM inferences were affected by the their domain knowledge more than those of the human reasoners. 

\textbf{Future Work.} There are numerous avenues for future research. 
% mention limitations of model fits and discuss more detailed analysis as one potential future avenue , if space permits
Here we compared human and LLM inferences on only one simple causal structure, whereas humans have been tested on causal networks with different topologies (e.g., forks, chain, etc.), causal relations (inhibitory vs. generative), integration functions (e.g., causes that combine conjunctively rather than independently), with more than three variables, and with continuous variables rather than binary ones. Besides the simple causal inferences examined here, there is a wealth of data on how humans intervene on causal systems, make causal attributions in cases of actual causation, and learn causal systems from observed data. Regarding LLMs, a deeper analysis of the effects of domain knowledge on their inferences is warranted as such knowledge can affect both independence (via inferred causal connections between the collider's causes) and explaining away \cite<via treating the two causal relations as interactive rather than independent;>[]{cruz2020explainingaway, morris1995one}.
It is also important to better understand how their inferences are affected by factors such as the temperature parameter.

% updated with new data
Overall, the tested LLMs largely engaged appropriately with the same complex prompts used in research on  human causal reasoning. GPT-4o's responses aligned most closely with normative inferences, with Gemini exhibiting similar performance. Claude, while slightly less normatively aligned than the former two,
%GPT4o and Gemini,  
more closely mirrored human reasoning patterns than GPT-4o. Notably, Gemini achieved both high normative consistency and the highest correlation with humans ($r_s = .763$).
GPT-3.5 deviated markedly from both with the exception of the predictive inference tasks.
% HD: can we really claim this? I softened it by inserting "lightly"
%Rather than treating the task abstractly, LLMs likely drew inferences based on their domain knowledge.

\bibliographystyle{apacite}

\setlength{\bibleftmargin}{.125in}
\setlength{\bibindent}{-\bibleftmargin}

\bibliography{submission_2025.bib}

%\newpage
%%%%%%%%%%%%%%%%%%%%%%%%%%%%%%%%%%%%%%%%%%%%%%%%%%%%%%%%%%%%%%%%%%%%%%%%%%%%%%%%%%%%%%%%%%%% APPENDIX
%%%%%%%%%%%%%%%%%%%%%%%%%%%%%%%%%%%%%%%%%%%%%%%%%%%%%%%%%%%%%%%%%%%%%%%%%%%
%\input{maybe_someday_useful_text/appendix.tex}  

\end{document}